\documentclass[runningheads]{llncs}
\usepackage{graphicx}
\usepackage{tikz}
\usepackage{comment}
\usepackage{amsmath,amssymb} 
\usepackage{color}
\usepackage{multirow}
\usepackage[accsupp]{axessibility}

\usepackage{threeparttable}
\usepackage{booktabs}
\usepackage{algpseudocode}
\usepackage{algorithm}
\usepackage{algorithmicx}
\usepackage{setspace}
\usepackage{stfloats}
\usepackage{colortbl}
\usepackage{marvosym}
\usepackage[colorlinks = true,
            linkcolor = red,
            urlcolor  = magenta,
            citecolor = green,
            anchorcolor = blue]{hyperref}
\begin{document}
\pagestyle{headings}
\mainmatter

\title{Decoupled Mixup for Generalized Visual Recognition}

\authorrunning{H. Liu et al.}
\titlerunning{Decoupled Mixup for Generalized Visual Recognition}

\author{Haozhe Liu\inst{\dagger \; 1}, Wentian Zhang\inst{\dagger \; 2}, Jinheng Xie\inst{\dagger \; 2}, Haoqian Wu\inst{3}, Bing Li\inst{\textrm{\Letter} \; 1}, \\ 
Ziqi Zhang\inst{4}, Yuexiang Li\inst{\textrm{\Letter} \; 2}, Yawen Huang\inst{2}, Bernard Ghanem\inst{1}, Yefeng Zheng\inst{2}}

\institute{
King Abdullah University of Science and Technology, Saudi Arabia \and
Jarvis Lab, Tencent, Shenzhen, China \and 
YouTu Lab, Tencent, Shenzhen, China \and
Tsinghua University, Shenzhen, China \\
\email{\{haozhe.liu;bing.li;bernard.ghanem\}@kaust.edu.sa; zhangwentianml@gmail.com; xiejinheng2020@email.szu.edu.cn}
\email{zq-zhang18@mails.tsinghua.edu.cn; \{linuswu;vicyxli;yawenhuang;yefengzheng\}@tencent.com}
}

\footnotetext{$\dagger$ Equal Contribution}
\footnotetext{This paper is accepeted by ECCV'22 Workshop (Causality in Vision)}
\maketitle

\begin{abstract}

Convolutional neural networks (CNN)  have demonstrated   remarkable performance, when the training and testing data  are from the same distribution. However, such trained CNN models often largely degrade on  testing data which is unseen and Out-Of-the-Distribution (OOD). To address this issue,   we propose a novel "Decoupled-Mixup" method  to train CNN models for OOD visual recognition. Different from previous work  combining pairs of images homogeneously,  our method decouples each image into discriminative  and noise-prone regions, and then  heterogeneously combine these regions of image pairs to train CNN models. Since the observation is that noise-prone regions such as textural and clutter background are adverse  to the generalization ability of CNN models during training,  we enhance features from discriminative regions and suppress  noise-prone ones when combining an image pair. To further improves the generalization ability of trained models, we propose to disentangle discriminative  and  noise-prone regions  in frequency-based  and context-based fashions.
Experiment results show the high generalization performance of our method   on testing data that are  composed of unseen contexts, where our method achieves  85.76\%  top-1 accuracy in Track-1 and 79.92\% in Track-2  in NICO Challenge.
The source code is available at \url{https://github.com/HaozheLiu-ST/NICOChallenge-OOD-Classification}. 
\end{abstract}

\section{Introduction}

Convolutional neural networks (CNN) have been successfully applied in various tasks such as visual recognition and image generation. However, the learned CNN models are vulnerable to the samples which are unseen and  Out-Of-Distribution (OOD) \cite{liu2021group,zhang2022nico++,liu2022robust}. To address this issue,  research efforts have been  devoted to data augmentation and regularization, which have shown   promising achievements.

Zhang \emph{et al.} \cite{zhang2017mixup} propose a  data augmentation method named  \textit{Mixup} which mixes image pairs and their corresponding labels to form  smooth annotations for training models. Mixup can be regarded as a locally linear out-of-manifold regularization \cite{guo2019mixup}, relating to the boundary of the adversarial robust training \cite{zhang2020does}. Hence, this simple technique has been shown to substantially facilitate both the model robustness and generalization. 
Following this direction, many variants have been proposed to explore the form of interpolation. Manifold Mixup \cite{verma2019manifold} generalizes Mixup to the feature space. Guo \emph{et al.} \cite{guo2019mixup} proposed an adaptive Mixup by reducing the misleading random generation. Cutmix is then proposed by Yun \emph{et al.} \cite{yun2019cutmix}, which introduces region-based interpolation between images to replace global mixing. By adopting the region-based mixing like Cutmix, Kim \emph{et al.} \cite{kim2020puzzle} proposed Puzzle Mix to generate the virtual sample by utilizing saliency information from each input. Liu \emph{et al.} \cite{liu2022robust} proposed to regard mixing-based data augmentation as a dynamic feature aggregation method, which can obtain a compact feature space with strong robustness against the adversarial attacks. 
More recently, Hong \emph{et al.} \cite{hong2021stylemix} proposed styleMix to separate content and style for enhanced data augmentation. As a contemporary work similar to the styleMix, Zhou \emph{et al.} \cite{zhou2021domain} proposed Mixstyle to mix the style in the bottom layers of a deep model within un-mixed label. By implicitly shuffling the style information, Mixstyle can improve model generalization and achieve the satisfactory OOD visual recognition performance. Despite of the gradual progress, Mixstyle and StyleMix should work based on AdaIN \cite{huang2017arbitrary} to disentangle style information, which requires the feature maps as input. However, based on the empirical study \cite{gatys2016image}, the style information is sensitive to the depth of the layer and the network architecture, which limits their potential for practical applications.

In this paper, inspired by Mixup and StyleMix, we propose a novel method named Decoupled-Mixup to combine image pairs for training CNN models. Our insight is that not all image regions benefit OOD visual recognition,  where noise-prone regions such as textural and clutter background are often adverse to the generalization of CNN
models during training.  Yet, previous work Mixup treats all image regions equally to combine a pair of images. Differently, we propose to  decouple each image into discriminative and noise-prone regions, and suppress noise-prone region during image combination, such that the CNN model pays more attention to discriminative regions  during training. 
In particular, we propose  an universal form based on Mixup, where  StyleMix can be regarded as a special case of Decoupled-Mixup in feature space.
Furthermore, by extending our Decoupled-Mixup to context and frequency domains respectively, we propose Context-aware  Decoupled-Mixup (CD-Mixup) and Frequency-aware Decoupled Mixup (FD-Mixup) to capture  discriminative regions
and  noise-prone ones using saliency  and the texture information, and suppress noise-prone regions in the image pair combination.  
By such heterogeneous combination,  our method trains the CNN model to emphasize more informative regions, which improves the  generalization ability of the trained model.

In summary, our contribution of  this paper is three-fold:
\begin{itemize}
   \item We propose a novel method to train CNN models for OOD visual recognition. Our method suppresses  noise-prone regions when combining image pairs for training, such that  the trained CNN model emphasizes  discriminative image regions, which  improves its generalization ability. 
    \item  Our CD-Mixup and FD-Mixup modules effectively decouple each image into discriminative and noise-prone regions by separately exploiting context and texture domains, which does not require extra object/instance-level  annotations.
  \item  Experiment results show that our method achieves superior performance and better generalization ability on testing data composed of unseen contexts, compared with state-of-the-art Mixup-based methods. 
\end{itemize}

\section{Related Works}

\subsubsection{OOD Generalization}

OOD generalization considers the generalization capabilities to the unseen distribution in the real scenarios of deep models trained with limited data. 
Recently, OOD generalization has been introduced in many visual applications \cite{liu2021fingerprint,zhang2021face,liu2021one,zhang2022towards,zhang2021deep}. In general, the unseen domains of OOD samples incur great confusion to the deep models in visual recognition. To address such issue, domain generalization methods are proposed to train models only from accessible domains and make models generalize well on unseen domains. Several works \cite{piratla2020efficient,zhang2021deep,zhang2022towa} propose to obtain the domain-invariant features across source domains and inhibit their
negative effect, leading to better generalization ability under unseen domain. 
Another simple but effective domain generalization method is to enlarge the data space with data augmentation and regularization of accessible source domains \cite{zhang2017mixup,zhou2020deep,zhou2020learning}. Following this direction, we further decouple and suppress the noise-prone regions (e.g. background and texture information) from source domains to improve OOD generalization of deep models. 

\subsubsection{Self/weakly Supervised Segmentation}
A series of methods~\cite{ornet,clims,xie2022c2am} demonstrate a good ability to segment objects of interest out of the complex backgrounds in Self/weakly supervised manners. However, in 
 the absence of pixel-level annotations, spurious correlations will result in the incorrect segmentation of class-related backgrounds. To handle this problem, CLIMS~\cite{clims} propose a language-image matching-based suppression and C$^2$AM~\cite{xie2022c2am} propose contrastive learning of foreground-background discrimination. The aforementioned methods can serve as the disentanglement function in the proposed context-aware Decoupled-Mixup.

\section{Method}

\begin{figure}[h]
  \centering
   \includegraphics[width=1.0\linewidth]{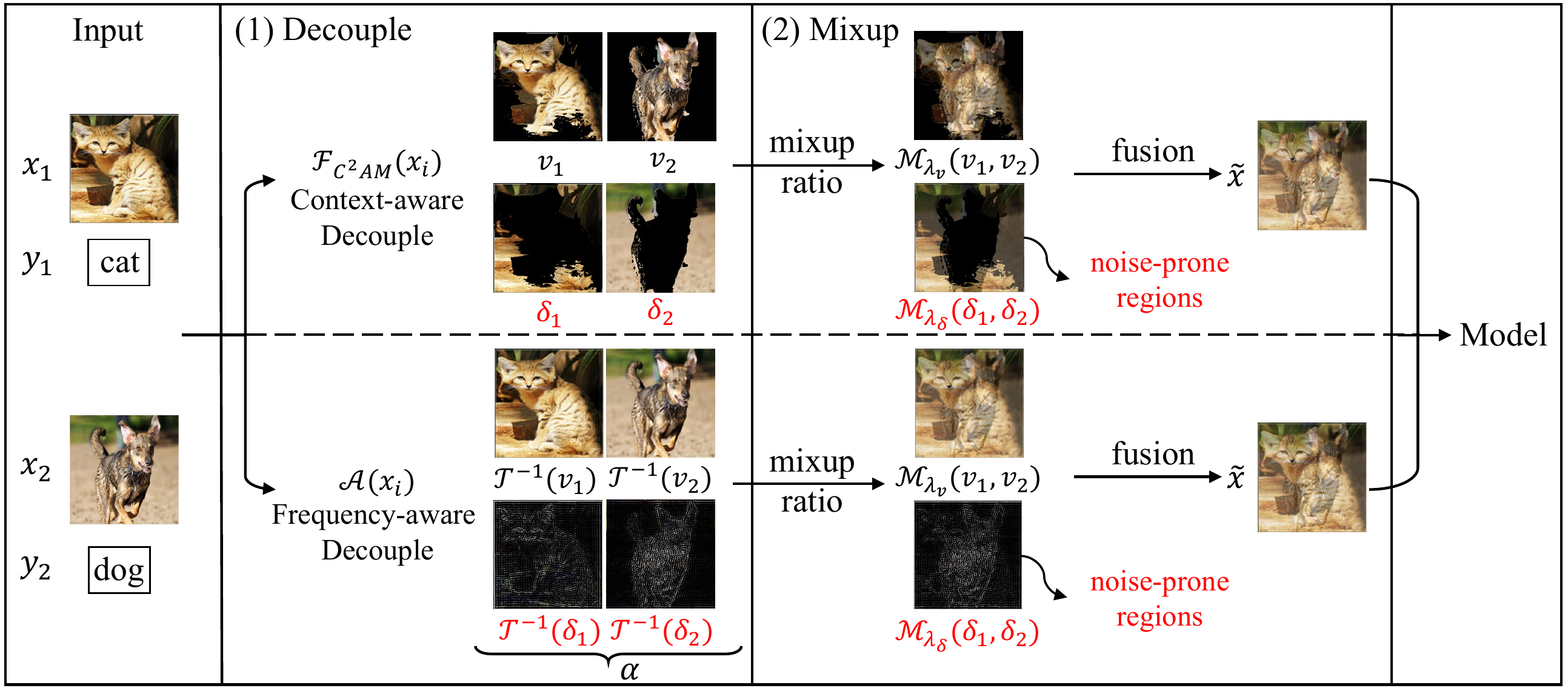}
   \caption{The overview of the proposed method. Given an input image pair and  and their annotations,  our method first decouples discriminative  and noise-prone regions for each image in the  context and frequency domains, respectively. Then, our method separately fuses  the discriminative  and noise-prone regions as well as the annotations, such that fused images encourage  CNN models to pay attention to  discriminative regions and neglect noise-prone ones  during training.}
   \label{fig:pipeline}
\end{figure}

We propose  Decoupled-Mixup to train CNN models for OOD visual
recognition.  Different from previous Mixup method combining pairs of images homogeneously,  we  propose to decouple each image into discriminative  and noise-prone regions, and then  heterogeneously combine these regions of image pairs. As shown in Fig. \ref{fig:pipeline}, our method decouples discriminative and noise-prone regions for each
image in different domains, respectively.  Then, our method separately fuses the discriminative and noise-prone regions  with different ratios.    The annotation  labels are also mixed.  By such heterogeneously combinations,   we argue that our method 
 tends to  construct discriminative visual  patterns for virtual training samples,  while  reducing  new noisy visual patterns. 
As a result, the fused  virtual samples encourage CNN models to pay attention to discriminative regions during training, which can improve  the generalizability of the trained CNN model.

\subsubsection{Revisiting Mixup.} Mixup \cite{zhang2017mixup} performs as a very active research line, due to its simple implementation and effectiveness. In essence, the model is trained by the convex combinations of pairs of examples and the corresponding labels. By adopting cross-entropy learning objective as an example, Mixup can be formulated as: 
\begin{align}
    \tilde{x} &= \mathcal{M}_{\lambda}(x_i,x_j) = \lambda  x_i + (1-\lambda) x_j, \\
    \tilde{y} &= \mathcal{M}_{\lambda}(y_i,y_j), \\
    \mathcal{L}_{\text{Mixup}} &=  - \sum \tilde{y}log(\mathbf{D}(\tilde{x})),
\end{align}
where $\mathcal{M}_{\lambda}(x_i,x_j)$ is the convex combination of the samples $x_i$ and $x_j$ with the corresponding label $y_i$ and $y_j$, respectively. The learning objective $\mathcal{L}_{\text{Mixup}}$ of Mixup can be regarded as the cross-entropy between $\tilde{y}$ and the prediction $\mathbf{D}(\tilde{x})$. Mixup can be interpolated as a form of data augmentation that drives the model $\mathbf{D}(\cdot)$ to behave linearly in-between training samples.

\subsubsection{Decoupled-Mixup.}
 Unlike typical supervised learning with sufficient training data, the training data of OOD visual recognition is limited, \textit{i.e.}, the testing samples are unseen and their distributions are  different  from the training data. We argue that noise-prone regions such as undesirable style information,  textural and clutter background are adverse  to the generalization of CNN models during training. Yet,  Mixup  directly mixes two images, which is the main bottleneck  for OOD visual recognition.
 In other words,  the interpolation hyperparameter $\lambda$ is randomly decided in the original Mixup. As a result, the trained model cannot determine whether a visual pattern is useful for domain generalization or not. Based on such observations, we generalize Mixup to a pipeline, which can be explained as `first decouple then mixing and suppressing', and is formulated  as:
\begin{align}
  v_i,\delta_i&=\mathcal{F}(x_i) \quad   \text{and} \quad v_j,\delta_j = \mathcal{F}(x_j) , \\
  \label{eq:mixing}
  \tilde{x} &=  \mathcal{M}_{\lambda_v}(v_i,v_j) + \mathcal{M}_{\lambda_\delta}(\delta_i,\delta_j), \\
  \label{eq:label_mix}
  \tilde{y} &= \alpha  \mathcal{M}_{\lambda_v}(y_i,y_j) + (1-\alpha) \mathcal{M}_{\delta}(y_i,y_j)
\end{align}
where $\mathcal{F}(\cdot)$ refers to the disentanglement function, which can separate common pattern $v$ from noise-prone regions $\delta$, $\alpha$ rectifies the weights for common patterns and  noise-prone regions. Note that $\lambda$ is an important parameter for interpolation, and $v$ and $y$ share the same $\lambda$, \emph{i.e.} $\lambda_v$. Below, we provide three kinds of disentanglement function  $\mathcal{F}(\cdot)$, including style-, context- and frequency-aware Decoupled-Mixup, where Style-based Decoupled-Mixup, (also called MixStyle or StyleMix) is proposed in \cite{zhou2021domain}. {Style-based Decoupled-Mixup can be regarded as a special case of  Decoupled-Mixup using Style information.}

\subsection{Style-based Decoupled-Mixup}
By following the work \cite{zhou2021domain,hong2021stylemix}, adaptive instance normalization (AdaIN) is adopted for disentangling the style and content information, which can be formed as
\begin{align}
    \text{AdaIN}(u_i,u_j) = \sigma(u_j) (\frac{u_i-\mu(u_i)}{\mu(u_i)}) + \mu(u_j)
\end{align}
where $u_i$ and $u_j$ refer to the feature map extracted from $x_i$ and $x_j$ respectively. The mean $\mu(\cdot)$ and variance $\sigma(\cdot)$ are calculated among spatial dimension independently for different channels. Given two samples, $x_i$ and $x_j$ as input, we can obtain four mixed features 
\begin{align}
    \begin{split}
     u_{ii} = \text{AdaIN}(u_i,u_i), \\
    u_{jj} = \text{AdaIN}(u_i,u_i), \\
    u_{ij} = \text{AdaIN}(u_i,u_j), \\
    u_{ji} = \text{AdaIN}(u_j,u_i), \\
    \end{split}
\end{align}
where $\{u_{ii},u_{jj},u_{ij},u_{ji}\}$ is the set by separately combining the content and style information. For example, $u_{ij}$ can be regarded as the combination with content information from $u_i$ and the style information from $u_j$. In the work \cite{hong2021stylemix}, the common pattern can be regarded as:
\begin{align}
   \mathcal{M}_{\lambda_v}(v_i,v_j) =  \underbrace{tu_{ii} + (\lambda_{v} - t)u_{ij}}_{\text{The part of } v_i} +  \underbrace{(1-\lambda_{v}) u_{jj}}_{\text{The part of }v_j},
\end{align}
and the noise-prone regions can be regarded as: 
\begin{align}
    \mathcal{M}_{\lambda_\delta}(\delta_i,\delta_j) = \delta_i = \delta_j = (u_{ji} - u_{jj}).
\end{align}
Then the annotation can be defined as:
\begin{align}
    \tilde{y} &= \alpha  \mathcal{M}_{\lambda_v}(y_i,y_j) + (1-\alpha) \mathcal{M}_{\lambda_\delta}(y_i,y_j)
\end{align}
which is identical with the learning objective reported in \cite{hong2021stylemix}. In other words, when $\mathcal{F}(\cdot)$ disentangles the style information from content information, Decoupled-Mixup is the same as StyleMix \cite{hong2021stylemix}.

\begin{figure}[h]
  \centering
   \includegraphics[width=1.0\linewidth]{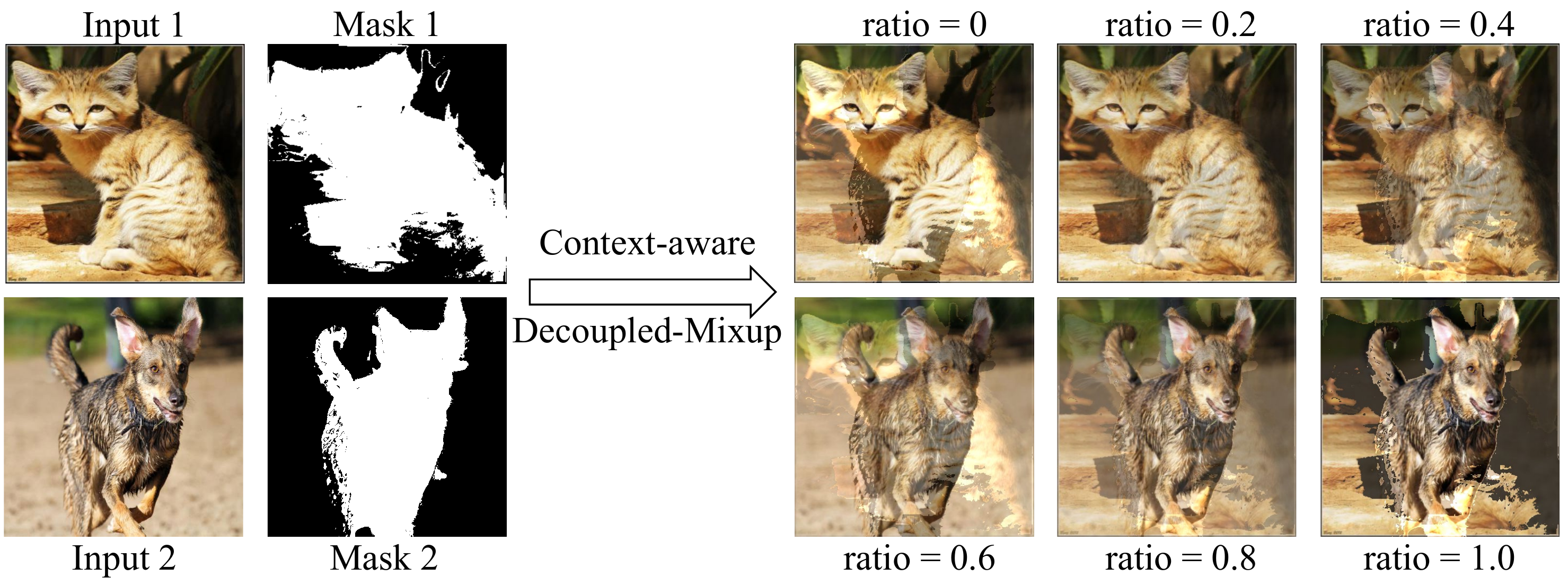}
   \caption{The visualization of context-aware Decoupled-Mixup method using two inputs from NICO Challenge. Their masks of foreground and background are extracted by C$^2$AM \cite{xie2022c2am}. Mixed images are showed when applying different foreground mixing ratios. }
   \label{fig:context_based}
\end{figure}

\subsection{Context-aware Decoupled-Mixup}
In OOD setting, the background of the images are quite different, which can be regarded as the noise-prone regions for visual recognition. In order to mitigate the influence caused by background, we propose a context-aware Decoupled-Mixup, where the disentanglement function $\mathcal{F}(\cdot)$ is designed to separate the foreground and background. In particular, we adopt an unsupervised method to extract the saliency region. In this paper, C$^2$AM \cite{xie2022c2am} is used as an example, which is based on contrastive learning to disentangle foreground and background. C$^2$AM can be formed as:
\begin{align}
    \mathcal{F}_{\text{C$^2$AM}}(x_i) = v_i, \delta_i, 
\end{align}
where $v_i$ refers to the foreground and $\delta_i$ is the background for $x_i$. Note that, C$^2$AM is an unsupervised method, which only depends on the pre-defined contrastive learning paradigm and thus prevents from any extra information. Then, to alleviate the influence from $\delta$, Decoupled-Mixup separately mixes the foreground and background by following Eq. (\ref{eq:mixing}) (as shown in Fig. \ref{fig:context_based}), and the annotation is fused by following Eq. (\ref{eq:label_mix}). When $\alpha=1$, background is mixed randomly, and foreground is mixed by following the ratios used in $y$. It can be explained as suppressing the extraction of background. 

\begin{figure}[t]
  \centering
   \includegraphics[width=0.85\linewidth]{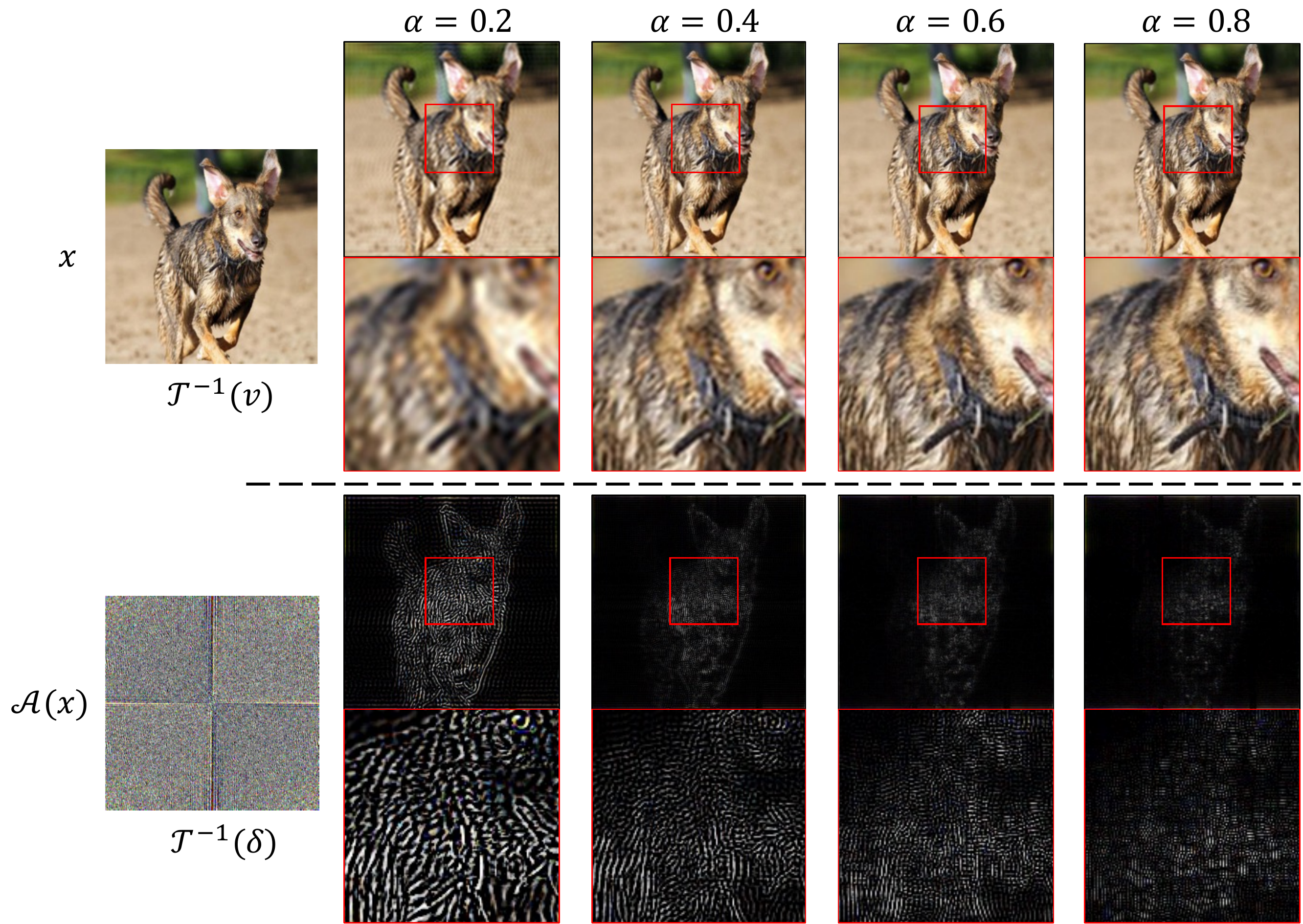}
   \caption{The example of Fourier transformation and its inverse transformation in NICO Challenge. $\mathcal{T}^{-1}(v)$ shows the low frequency part of images and $\mathcal{T}^{-1}(\delta)$ obtains high frequency parts. These two parts are well separated. 
   }
   \label{fig:fourier_1}
\end{figure}
\subsection{Frequency-aware Decoupled-Mixup}
\begin{figure*}[h]
  \centering
   \includegraphics[width=0.98\linewidth]{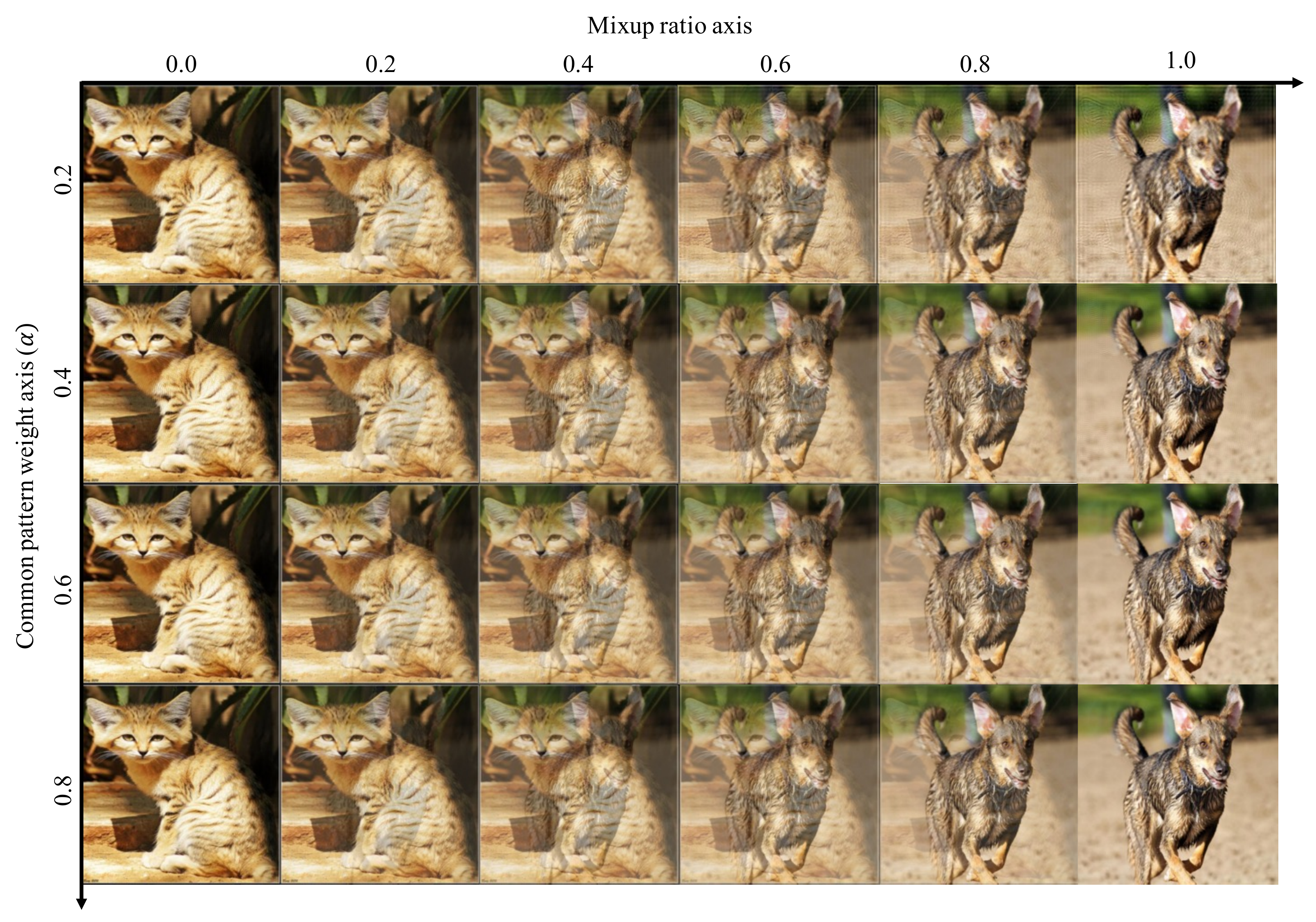}
   \caption{A grid visualization of mixed images using Frequency-aware Decoupled-Mixup method by adjusting the common pattern weight and mixup ratio. The low frequency part of each image are mixed, and the high frequency parts are well suppressed. }
   \label{fig:fourier_2}
\end{figure*}

In addition to the background, the textural region can lead to noisy information for training CNN models. In order to learn feature from discriminative regions, we generalize Mixup into frequency field, since the high-frequency component can be regarded as the texture information to some extent. In other words, the high-frequency component is the noise-prone regions,  and the low-frequency component refers to common patterns. 
In this paper, we adopt Fourier transformation $\mathcal{T}(\cdot)$ to capture the frequency information. Fourier transformation $\mathcal{T}(\cdot)$ is formulated as: 
\begin{align}
    \mathcal{T}(x)(u,v) = \sum_{h=0}^{H} \sum_{w=0}^{W} x(h,w) e ^{-j2\pi(\frac{h}{H}u+\frac{w}{W}v)},
\end{align}
where $\mathcal{T}(\cdot)$ can be easily implemented by Pytorch Library \cite{paszke2019pytorch}. To detect common patterns, we calculate the amplitude $\mathcal{A}(\cdot)$ of $x$ as follows: 
\begin{align}
    \mathcal{A}(x)(u,v) = [R^2(x)(u,v) + I^2(x)(u,v)]^{\frac{1}{2}},
\end{align}
where $R(x)$ and $I(x)$ refers to the real and imaginary part of $\mathcal{T}(x)$, respectively. 
The common pattern $v_i$ can be defined as $\mathcal{A}(x_i)(u<\alpha*H,v<\alpha*W)$ and the $\delta_i$ refers to the complement part of $\mathcal{A}(x_i)$, as shown in Fig. \ref{fig:fourier_1}.  Since the mixing occurs in frequency field, we should reverse mixed $\mathcal{A}(x_i)$ to image space, which can be expressed as:
\begin{align}
    \tilde{x} = \mathcal{T}^{-1} [\mathcal{M}_{\lambda_v}(v_i,v_j) + \mathcal{M}_{\lambda_\delta}(\delta_i,\delta_j)],
\end{align}
where $\mathcal{T}^{-1} (\cdot)$ is the  inverse Fourier transformation.  The mixed images are shown in Fig. \ref{fig:fourier_2}. Note that the annotation of $\tilde{x}$ is calculated using the ratio applied in $\mathcal{M}_{\lambda_v}(v_i,v_j)$.

\subsection{Decoupled-Mixup for OOD Visual Recognition}
Decoupled-Mixup can be easily combined with other methods or common tricks friendly, due to its simple implementation and effectiveness. Hence, we present the tricks and the other methods adopted for NICO Challenge.

\noindent
\subsubsection{Heuristic Data Augmentation.}
To further improve the generalization of the trained model, our method adopts extra data augmentation approaches. Based on the empirical study, heuristic data augmentation, \emph{i.e.} AugMix \cite{hendrycks2020augmix}, is applied in this paper. AugMix is characterized by its utilization of simple augmentation operations, where the augmentation operations are sampled stochastically and layered to produce a high diversity of augmented images. Note that, in order to alleviate the extra computation cost, we do not use Jensen-Shannon divergence as a consistency loss, which is reported in the conventional work \cite{hendrycks2020augmix}.

\noindent
\subsubsection{Self-supervised Learning.}
In NICO challenge, no extra data can be used for data training. In order to find a reasonable weight initialization, we adopt MoCo-V2 \cite{chen2020improved} to determine the weight initialization before vanillia model training.

\noindent
\subsubsection{Student-Teacher Regularization (S-TR).}
Following the work \cite{xu2021fourier}, we adopt teacher-student regularization to train the model. Specifically, a dual-formed consistency loss, called student-teacher regularization, is further introduced between the predictions induced from original and Fourier-augmented images.

\noindent
\subsubsection{Curriculum Learning (CL).}
Since OOD setting is a hard case for the deep models to learn, we gradually design the learning objective for training. We split training strategy into three stage. In the first stage, the image size is set as 224 $\times$ 224, and trained with normal data without mixing.  Then, in the second stage, we adopt Decoupled-Mixup to train the model with 448 $\times$ 448 image size. Finally, we introduce Student-Teacher regularization to search for the optimal model with 512 $\times$ 512 image size.

\noindent
\subsubsection{Model Ensembles (ME).}
In this paper, we ensemble several famous backbones, including ResNet-34 \cite{he2016deep}, WideResNet-50 \cite{zagoruyko2016wide} and DenseNet-121 \cite{huang2017densely} to finish the prediction. In addition to the architecture-wise ensemble, we also conduct the hyperparameter-wise ensemble. When the model is trained via curriculum learning, we use different hyperparameters, including learning rate, $\alpha$, etc., to fine-tune the trained model, and then average their outputs. Finally, the prediction is calculated among different architectures and hyperparameters by simply averaging their outputs.

\noindent
\subsubsection{Heuristic Augmentation based Inference.}
Since the predictions using different data augmentations respond differently to a distribution shift, we fuse these predictions into a final prediction. In particular, given an input image, we first employ AugMix to augment the image multiple times, and then average the prediction results of all these augmented images as the final result.

\section{Experiment}
We evaluate our Decoupled-Mixup method on Track-1 and Track-2 in NICO Challenge. Firstly, We verify our method on different models. Then, the  method is compared with other Mixup methods and different weight settings. Finally, we provide experimental results of our Decoupled-Mixup method using different training tricks. 

\subsection{Experiment Settings}

We apply the same experimental settings in Track-1 and Track-2. The batch size is 128, and the training epochs are 200. The learning rate starts at 0.1 and is decayed through a cosine annealing schedule. We used ResNet-34 \cite{he2016deep}, WideResNet-50 \cite{zagoruyko2016wide} and DenseNet-121 \cite{huang2017densely} as classifiers. Our Context-aware Decoupled-Mixup  follows the settings of C$^2$AM \cite{xie2022c2am} to obtain the image masks. In Frequency-aware Decoupled-Mixup, the common pattern weight $\alpha$ is set to 0.2 and 0.6 in Track1 and Track-2, respectively. To further show the generalization ability of our method,  we use the data of Track-1 for training and validate the trained model on the data of Track-2, and vice versa (\emph{i.e.} cross-track validation), besides  public test setting.
Note that we do not adopt this cross-track setting for Phase I competition to ensure the fair competition.

\subsection{Classification Results}

\begin{table}[h]
\centering
\caption{Top-1 accuracy (\%) of different models with or without our Decoupled-Mixup method.}
\resizebox{0.7\textwidth}{!}{
\begin{tabular}{l|cc|cc}
\hline
\multirow{2}{*}{Model} & \multicolumn{2}{c|}{Cross-track Validation} & \multicolumn{2}{c}{Public Test Set} \\ \cline{2-5} & Track1   & Track2  & Track1  & Track2      \\ \hline
ResNet-34                           & 85.80                 & 78.54                 & -                        & 68.54               \\
ResNet-34 w/ Ours                    & 88.98                 & \textbf{84.84}                 & \textbf{81.81}                & 74.26                \\ \hline
DenseNet-121                        & 85.48                 & 82.56                 & -                        & 73.46               \\
DenseNet-121 w/ Ours                 & 88.22                 & 83.89                 & 81.75               & 74.13               \\ \hline
WideResNet-50                        & 85.36                 & 78.48                 & -                       & 68.39               \\
WideResNet-50 w/ Ours                 & \textbf{89.83}                 & 84.41                 & 80.99               & \textbf{74.29}              \\ \hline
\end{tabular}
}
\label{tab:1}
\end{table}

\begin{table}[h]
\centering
\caption{Top-1 accuracy (\%) of  our Decoupled-Mixup method using different values of  common pattern weight $\alpha$, where the backbone is ResNet-34.}
\setlength\tabcolsep{11pt}
\resizebox{0.65\textwidth}{!}{
\begin{tabular}{l|cc|cc}
\hline
\multirow{2}{*}{$\alpha$} & \multicolumn{2}{c|}{Cross-track Validation} & \multicolumn{2}{c}{Public Test Set} \\ \cline{2-5} & Track1   & Track2  & Track1  & Track2      \\ \hline
0.1 & 88.95 & 83.56 &-  & 72.95 \\
0.2   & \textbf{89.54} & 83.80   &-  & 72.96 \\
0.4   & 89.05 & 84.26 &-  & 73.31 \\
0.6   & 88.98 & \textbf{84.84} &-  & 73.08 \\
0.8   & 89.36 & 84.56 &-  & \textbf{73.85} \\ \hline
\end{tabular}
}
\label{tab:2}
\end{table}

\textbf{Comparison.} To show the effectiveness of our method,  we use our method and state-of-the-art Mixup methods to train  WideResNet-50 Model, respectively.  As shown in Table \ref{tab:3},   our Decoupled-Mixup method  achieves the best classification performance, compared with Mixstyle, Mixup, and CutMix methods.
 Mixstyle methods and other feature space based methods are thus hard to be effective in the training phase, 
 since the WideResNet-50 model is trained from scratch and no prior knowledge can be provided into latent layers of WideResNet-50 Model.

In addition,  we use our method to train various wide-used backbones: ResNet-34, DenseNet-121 and WideResNet-50. As shown in Table \ref{tab:1},  our method improves the performance of all these models in both cross-track validation and public test set. 

\textbf{Parameter tuning.} To further discover the noise-prone regions in frequency domain, Table \ref{tab:2} compares the classification results of ResNet-34 using different values of common pattern weight ($\alpha$). When  $\alpha=$  0.2 in Track-1 and 0.6 in Track-2, more noise-prone regions can be suppressed.

\begin{table}[h]
\centering
\caption{Top-1 accuracy (\%) of  different Mixup methods using WideResNet-50 model.}
\resizebox{0.6\textwidth}{!}{
\begin{tabular}{l|cc|cc}
\hline
\multirow{2}{*}{Mixup Method} & \multicolumn{2}{c|}{Cross-track Validation} & \multicolumn{2}{c}{Public Test Set} \\ \cline{2-5} & Track1   & Track2  & Track1  & Track2      \\ \hline
Baseline                        & 85.36                 & 78.48                 &-                         & 68.39               \\
Mixstyle \cite{zhou2021domain}               & 58.71                 & 63.99                  & -                       & -                      \\
Mixup \cite{zhang2017mixup}                 & 86.76                 & 80.52                 & -                       & -               \\
CutMix \cite{yun2019cutmix}                & 88.02                 & 77.31                  & -                      & -                       \\
Ours                 & \textbf{89.83}                 & \textbf{84.41}                 & 80.99               & 74.29 \\ \hline
\end{tabular}
}
\label{tab:3}
\end{table}


\textbf{Effect of training tricks.} 
We conduct experiments to investigate the effect of our training tricks  on our method. As shown in Table \ref{tab:4} shows the performance of our method using different tricks, where  all models are pretrained by MoCo-V2 \cite{chen2020improved} for the weight initialization. Through  model ensembling, the proposed Decoupled-Mixup method reaches to the 85.76\% in Track-1 and 79.92\% in Track-2 in public test set.  

\begin{table}[h!]
\centering
\caption{Top-1 accuracy (\%)  of our Decoupled-Mixup method  using different tricks, where of the backbone is WideResNet-50.}
\setlength\tabcolsep{6pt}
\resizebox{1.0\textwidth}{!}{
\begin{tabular}{cc|ccc|cc|cc}
\hline
\multicolumn{2}{c|}{Method}& \multicolumn{3}{c|}{Tricks}& \multicolumn{2}{c|}{Cross-track Validation} & \multicolumn{2}{c}{Public Test Set}  \\\hline
CD-Mixup & FD-Mixup & CL & S-TR & ME* & Track1 & Track2   & Track1 & Track2  \\ \hline
$\checkmark$ & $\times$     &$\times$    &$\times$   &$\times$     & 87.79 & 81.60 & -         & -         \\
$\times$ &$\checkmark$      &$\times$    &$\times$   &$\times$      & 88.27 & 80.45 & -         & -         \\      
$\checkmark$&$\checkmark$      &$\checkmark$    &$\times$   &$\times$     & 89.97  & 84.74 & 82.13         &74.16         \\    
$\checkmark$&$\checkmark$      &$\checkmark$    &$\checkmark$   &$\times$     & 90.07 & 85.21 & 82.63 & 76.10 \\  
$\checkmark$&$\checkmark$     &$\checkmark$    &$\checkmark$   &$\checkmark$     &\textbf{91.13}        & \textbf{88.54}       & \textbf{85.76}     & \textbf{79.92}  \\ \hline  
\end{tabular}
}
\label{tab:4}
\begin{tablenotes}
\footnotesize
\item[*] \small{* ResNet-34, DenseNet-121 and WideResNet-50 models trained by the proposed Decoupled-Mixup method are ensembled to finish the prediction.}
\end{tablenotes}
\end{table}


\section{Conclusion}
In this paper, we propose a novel method to  train CNN models for OOD visual recognition. Our method proposes to  decouples an image into discriminative and  noise-prone regions, and suppresses  noise-prone regions when combining image pairs for training,
 CD-Mixup and FD-Mixup are proposed to  decouple each image into discriminative and noise-prone regions in context and texture domains,  by exploiting saliency and textual information. By  suppressing   noise-prone regions in the image combination,  our method effectively enforce  the trained CNN model to emphasize  discriminative image regions, which  improves its generalization ability.  Extensive experiments show the superior performance of the proposed method, which reaches to 4th/39 in the Track-2 of NICO Challenge in the final ranking.

\subsubsection{Acknowledgements} 
We would like to thank the efforts of the NICO Challenge officials, who are committed to maintaining the fairness and openness of the competition. We also could not have undertaken this paper without efforts of every authors. 
This work was supported in part by the Key-Area Research and Development Program of Guangdong Province, China (No. 2018B010111001), National Key R\&D Program of China (2018YFC2000702), in part by the Scientific and Technical Innovation 2030-``New Generation Artificial Intelligence" Project (No. 2020AAA0104100) and  in part by the
King Abdullah University of Science and Technology
(KAUST) Office of Sponsored Research through the Visual
Computing Center (VCC) funding.

\bibliographystyle{splncs04}
\bibliography{cite}

\begin{thebibliography}{10}
\providecommand{\url}[1]{\texttt{#1}}
\providecommand{\urlprefix}{URL }
\providecommand{\doi}[1]{https://doi.org/#1}

\bibitem{chen2020improved}
Chen, X., Fan, H., Girshick, R., He, K.: Improved baselines with momentum
  contrastive learning. arXiv preprint arXiv:2003.04297  (2020)

\bibitem{gatys2016image}
Gatys, L.A., Ecker, A.S., Bethge, M.: Image style transfer using convolutional
  neural networks. In: Proceedings of the IEEE conference on computer vision
  and pattern recognition. pp. 2414--2423 (2016)

\bibitem{guo2019mixup}
Guo, H., Mao, Y., Zhang, R.: Mixup as locally linear out-of-manifold
  regularization. In: Proceedings of the AAAI Conference on Artificial
  Intelligence. vol.~33, pp. 3714--3722 (2019)

\bibitem{he2016deep}
He, K., Zhang, X., Ren, S., Sun, J.: Deep residual learning for image
  recognition. In: Proceedings of the IEEE conference on computer vision and
  pattern recognition. pp. 770--778 (2016)

\bibitem{hendrycks2020augmix}
Hendrycks, D., Mu, N., Cubuk, E.D., Zoph, B., Gilmer, J., Lakshminarayanan, B.:
  {AugMix}: A simple data processing method to improve robustness and
  uncertainty. Proceedings of the International Conference on Learning
  Representations (ICLR)  (2020)

\bibitem{hong2021stylemix}
Hong, M., Choi, J., Kim, G.: Stylemix: Separating content and style for
  enhanced data augmentation. In: Proceedings of the IEEE/CVF Conference on
  Computer Vision and Pattern Recognition. pp. 14862--14870 (2021)

\bibitem{huang2017densely}
Huang, G., Liu, Z., Van Der~Maaten, L., Weinberger, K.Q.: Densely connected
  convolutional networks. In: Proceedings of the IEEE conference on computer
  vision and pattern recognition. pp. 4700--4708 (2017)

\bibitem{huang2017arbitrary}
Huang, X., Belongie, S.: Arbitrary style transfer in real-time with adaptive
  instance normalization. In: Proceedings of the IEEE international conference
  on computer vision. pp. 1501--1510 (2017)

\bibitem{kim2020puzzle}
Kim, J.H., Choo, W., Song, H.O.: Puzzle mix: Exploiting saliency and local
  statistics for optimal mixup. In: International Conference on Machine
  Learning. pp. 5275--5285. PMLR (2020)

\bibitem{liu2021one}
Liu, F., Liu, H., Zhang, W., Liu, G., Shen, L.: One-class fingerprint
  presentation attack detection using auto-encoder network. IEEE Transactions
  on Image Processing  \textbf{30},  2394--2407 (2021)

\bibitem{liu2022robust}
Liu, H., Ji, H., Li, Y., He, N., Wu, H., Liu, F., Shen, L., Zheng, Y.: Robust
  representation via dynamic feature aggregation. arXiv preprint
  arXiv:2205.07466  (2022)

\bibitem{liu2021group}
Liu, H., Wu, H., Xie, W., Liu, F., Shen, L.: Group-wise inhibition based
  feature regularization for robust classification. In: Proceedings of the
  IEEE/CVF International Conference on Computer Vision. pp. 478--486 (2021)

\bibitem{liu2021fingerprint}
Liu, H., Zhang, W., Liu, F., Wu, H., Shen, L.: Fingerprint presentation attack
  detector using global-local model. IEEE Transactions on Cybernetics  (2021)

\bibitem{paszke2019pytorch}
Paszke, A., Gross, S., Massa, F., Lerer, A., Bradbury, J., Chanan, G., Killeen,
  T., Lin, Z., Gimelshein, N., Antiga, L., et~al.: Pytorch: An imperative
  style, high-performance deep learning library. Advances in neural information
  processing systems  \textbf{32} (2019)

\bibitem{piratla2020efficient}
Piratla, V., Netrapalli, P., Sarawagi, S.: Efficient domain generalization via
  common-specific low-rank decomposition. In: International Conference on
  Machine Learning. pp. 7728--7738. PMLR (2020)

\bibitem{verma2019manifold}
Verma, V., Lamb, A., Beckham, C., Najafi, A., Mitliagkas, I., Lopez-Paz, D.,
  Bengio, Y.: Manifold mixup: Better representations by interpolating hidden
  states. In: International Conference on Machine Learning. pp. 6438--6447.
  PMLR (2019)

\bibitem{clims}
Xie, J., Hou, X., Ye, K., Shen, L.: {CLIMS}: Cross language image matching for
  weakly supervised semantic segmentation. In: Proceedings of the IEEE/CVF
  Conference on Computer Vision and Pattern Recognition (CVPR). pp. 4483--4492
  (June 2022)

\bibitem{ornet}
Xie, J., Luo, C., Zhu, X., Jin, Z., Lu, W., Shen, L.: Online refinement of
  low-level feature based activation map for weakly supervised object
  localization. In: Proceedings of the IEEE/CVF International Conference on
  Computer Vision (ICCV). pp. 132--141 (October 2021)

\bibitem{xie2022c2am}
Xie, J., Xiang, J., Chen, J., Hou, X., Zhao, X., Shen, L.: {C2AM}: Contrastive
  learning of class-agnostic activation map for weakly supervised object
  localization and semantic segmentation. In: Proceedings of the IEEE/CVF
  Conference on Computer Vision and Pattern Recognition. pp. 989--998 (2022)

\bibitem{xu2021fourier}
Xu, Q., Zhang, R., Zhang, Y., Wang, Y., Tian, Q.: A fourier-based framework for
  domain generalization. In: Proceedings of the IEEE/CVF Conference on Computer
  Vision and Pattern Recognition. pp. 14383--14392 (2021)

\bibitem{yun2019cutmix}
Yun, S., Han, D., Oh, S.J., Chun, S., Choe, J., Yoo, Y.: Cutmix: Regularization
  strategy to train strong classifiers with localizable features. In:
  Proceedings of the IEEE/CVF international conference on computer vision. pp.
  6023--6032 (2019)

\bibitem{zagoruyko2016wide}
Zagoruyko, S., Komodakis, N.: Wide residual networks. arXiv preprint
  arXiv:1605.07146  (2016)

\bibitem{zhang2017mixup}
Zhang, H., Cisse, M., Dauphin, Y.N., Lopez-Paz, D.: mixup: Beyond empirical
  risk minimization. arXiv preprint arXiv:1710.09412  (2017)

\bibitem{zhang2020does}
Zhang, L., Deng, Z., Kawaguchi, K., Ghorbani, A., Zou, J.: How does mixup help
  with robustness and generalization? ICLR  (2021)

\bibitem{zhang2021face}
Zhang, W., Liu, H., Liu, F., Ramachandra, R., Busch, C.: Frt-pad: Effective
  presentation attack detection driven by face related task. arXiv preprint
  arXiv:2111.11046  (2021)

\bibitem{zhang2021deep}
Zhang, X., Cui, P., Xu, R., Zhou, L., He, Y., Shen, Z.: Deep stable learning
  for out-of-distribution generalization. In: Proceedings of the IEEE/CVF
  Conference on Computer Vision and Pattern Recognition. pp. 5372--5382 (2021)

\bibitem{zhang2022towards}
Zhang, X., Xu, Z., Xu, R., Liu, J., Cui, P., Wan, W., Sun, C., Li, C.: Towards
  domain generalization in object detection. arXiv preprint arXiv:2203.14387
  (2022)

\bibitem{zhang2022nico++}
Zhang, X., Zhou, L., Xu, R., Cui, P., Shen, Z., Liu, H.: Nico++: Towards better
  benchmarking for domain generalization. arXiv preprint arXiv:2204.08040
  (2022)

\bibitem{zhang2022towa}
Zhang, X., Zhou, L., Xu, R., Cui, P., Shen, Z., Liu, H.: Towards unsupervised
  domain generalization. In: Proceedings of the IEEE/CVF Conference on Computer
  Vision and Pattern Recognition. pp. 4910--4920 (2022)

\bibitem{zhou2020deep}
Zhou, K., Yang, Y., Hospedales, T., Xiang, T.: Deep domain-adversarial image
  generation for domain generalisation. In: Proceedings of the AAAI Conference
  on Artificial Intelligence. vol.~34, pp. 13025--13032 (2020)

\bibitem{zhou2020learning}
Zhou, K., Yang, Y., Hospedales, T., Xiang, T.: Learning to generate novel
  domains for domain generalization. In: European conference on computer
  vision. pp. 561--578. Springer (2020)

\bibitem{zhou2021domain}
Zhou, K., Yang, Y., Qiao, Y., Xiang, T.: Domain generalization with mixstyle.
  ICLR  (2021)

\end{thebibliography}

\end{document}